\documentclass{article}
\usepackage{ijcai26}

\usepackage{times}
\usepackage{soul}
\usepackage{url}
\usepackage[hidelinks]{hyperref}
\usepackage[utf8]{inputenc}
\usepackage[small]{caption}
\usepackage{graphicx}
\usepackage{amsmath}
\usepackage{amsthm}
\usepackage{booktabs}
\usepackage{array}
\usepackage[section]{placeins}
\usepackage{subcaption}
\usepackage[switch]{lineno}
\usepackage{placeins}
\title{PazaBench: Large-Scale ASR Benchmark for Low-Resource Languages in Africa}

\author{
    Mercy Muchai \and
    Nick Mumero Mwangi \and
    Kevin Chege \and
    Samuel Chege Maina
    \affiliations
    Microsoft Research Africa, Nairobi
    \emails {\{mercymuchai, v-nickmwangi, v-kevinchege, samuelmaina\}@microsoft.com}
}
\begin{document}

\maketitle
\begin{abstract}
Low-resource languages remain underrepresented in ASR benchmarks, limiting the ability to reliably assess and compare model performance. We present a large-scale benchmark for automatic speech recognition (ASR) on low-resource languages in Africa. The benchmark harmonizes six public corpora into a unified, reproducible evaluation framework and evaluates 52 state-of-the-art ASR models spanning three model architectural paradigms across 39 African languages on diverse datasets. Models are evaluated using standard accuracy and efficiency metrics, word error rate (WER), character error rate (CER), and inverse real-time factor (RTFx). This benchmark introduces architecture-aware comparison that systematically analyzes both individual models and broader model families. In addition, we quantify linguistic variance across language families, revealing how these factors shape ASR performance across diverse African languages. We also provide a comprehensive accuracy and efficiency trade-off analysis directly relevant to deployment in low-resource settings. Together, the benchmark, evaluation pipeline, and results support transparent, reproducible, and deployment-aware evaluation of ASR models for low-resource languages.
\end{abstract}

\section{Introduction}\label{sec:introduction}

Recent progress in automatic speech recognition (ASR) has been driven by large-scale multilingual training, self-supervised representation learning, and the integration of speech with large language models, resulting in strong performance on widely used benchmarks for high-resource languages~\cite{Radford2023Whisper}. These advances have broadened the ASR landscape, spanning encoder-only and encoder--decoder architectures as well as emerging LLM-based models, while substantially improving cross-lingual transfer capabilities.

However, evaluation practices have not kept pace in low-resource settings: Languages with limited labeled data introduce challenges related to linguistic diversity, morphology, orthography, and deployment constraints that are not well reflected in the existing benchmarks. Many multilingual ASR evaluations focus on aggregate results or high-resource subsets, which limits insights into model behavior in data-scarce and/or deployment-constrained settings.

African languages illustrate this gap clearly. Despite their linguistic diversity and societal relevance, they remain underrepresented in many multilingual ASR benchmarks, and when included, results are often summarized using macro-averaged metrics that obscure language-specific behavior~\cite{pratap2023mms}. This not only limits inclusivity but also constrains our ability to capture meaningful variation in model performance, reducing opportunities to uncover insights that such diversity can provide. Prior work has shown that languages with similar data availability can exhibit markedly different performance, reflecting differences in linguistic structure and model inductive biases~\cite{Baevski2020Wav2Vec2,Conneau2021XLSR}. As a result, existing benchmarks leave critical questions unanswered on ASR performance for African languages.

At the same time, the ASR ecosystem has expanded along architectural lines. Encoder-only models emphasize inference efficiency through non-autoregressive decoding~\cite{Baevski2020Wav2Vec2,Xu2023TDT}, encoder-decoder models prioritize transcription accuracy and robustness~\cite{Radford2023Whisper}, and emerging LLM-based approaches integrate ASR into broader multimodal frameworks~\cite{Chu2024Qwen2Audio,Abouelenin2025Phi4Mini}. Ignoring these architectural differences or inference efficiency on low-resource languages can obscure the roles of model scale, training data, and design choices, reducing the interpretability and deployment relevance of benchmark results.

\paragraph{Contributions.} We present PazaBench\footnote{\url{https://huggingface.co/spaces/microsoft/paza-bench}}, the first large-scale unified benchmark evaluating 52 publicly available ASR models across 39 low and mid-resource African languages under a standardized evaluation protocol. This benchmark introduces architecture-aware analysis supported by rigorous non-parametric statistical testing. We further quantify linguistic variance and complexity that are substantial enough to render aggregate multilingual averages misleading. Finally, we provide a comprehensive accuracy--efficiency ASR trade-off analysis, offering actionable insights for deploying robust ASR systems for low-resource languages.

\section{Related Work}\label{sec:relatedworks}

Evaluation of ASR models has traditionally relied on standardized benchmark datasets and metrics such as WER and CER, enabling comparison across models and training approaches. These benchmarks have played a central role in tracking ASR progress under ideal conditions.

Several large-scale multilingual datasets support evaluation across languages, including Common Voice~\cite{Ardila2020CommonVoice} and MLS~\cite{Pratap2020MLSAL}. Additionally, the CMU Wilderness corpus~\cite{Black2019Wilderness} provided one of the earliest large-scale multilingual speech collections covering African languages. While these resources provide broad multilingual coverage, current benchmarks predominantly emphasize high-resource languages and offer limited representation of low-resource languages or those with substantial dialectal variation.

Efforts to build speech representations and benchmarks for underrepresented languages have gained momentum. Self-supervised models such as XLS-R~\cite{Babu2022XLSR} and cross-lingual wav2vec~\cite{Conneau2021XLSR} demonstrated that pretraining on diverse multilingual audio enables transfer to low-resource settings. More targeted efforts include AfriHuBERT~\cite{Alabi2024AfriHuBERT}, a self-supervised model pretrained specifically on African language speech, and mHuBERT-147~\cite{Boito2024mHuBERT}, a compact multilingual speech model covering 147 languages including several African languages.~\cite{Chen2024RobustSpeechRep} further scale self-supervised pretraining to thousands of languages, including many African languages with minimal labelled data. Recent LLM-based systems further expand the scope of ASR evaluation by integrating speech recognition with broader language understanding tasks. Models such as Phi-4 multimodal-instruct~\cite{Abouelenin2025Phi4Mini} and Omnilingual ASR~\cite{OmnilingualASR2025} with 1600+ language coverage demonstrate strong performance, placing new demands on benchmarks to support architecture-aware and efficiency-aware evaluation across diverse languages.

Benchmark suites with broader task and model coverage have also emerged. AudioBench~\cite{wang2024audiobench} extends evaluation beyond ASR to include speech translation, audio captioning, and audio question answering, supporting comparison across varying architectures including LLM-based models such as Qwen-Audio~\cite{Chu2024Qwen2Audio}. However, its ASR evaluation remains largely concentrated on English. The OpenASR leaderboard~\cite{srivastav2025openasrleaderboardreproducible} introduces standardized evaluation of accuracy and inference efficiency across hybrid, CTC, and transformer-based models, but continues to focus primarily on high-resource languages.

Dedicated benchmarks for non-European language families have also emerged. ML-SUPERB~\cite{Shi2023MLSUPERB} provides a standardized multilingual benchmark spanning diverse language families, while IndicSUPERB~\cite{Javed2023IndicSUPERB} specifically targets Indian languages. For African languages, SimbaBench~\cite{Elmadany2025SimbaBench} aggregates speech data across 61 African languages and provides a leaderboard for ASR and related tasks. However, these benchmarks do not compare architectural paradigms or jointly evaluate accuracy and inference efficiency.

This highlights a broader evaluation gap: while dedicated African language datasets, models and few benchmarks exist, none jointly evaluate architecture-aware accuracy and efficiency metrics, and language complexity across diverse ASR paradigms for low-resource African languages.

\section{Benchmark Construction and Methodology}\label{sec:methodology}
This benchmark focuses on African languages with heterogeneous scripts, morphology, and data availability, while jointly evaluating transcription accuracy and efficiency under controlled conditions.

\subsection{Languages and Dataset Groups}

We construct a unified evaluation benchmark covering 39 African languages by aggregating six publicly available corpora:
AfriVoices-KE~\cite{wanzare2026afrivoiceske},  African Next Voices South Africa~\cite{marivate2025swivuriso0}, ALFFA~\cite{gauthier2016collect}, Google FLEURS~\cite{fleurs2022arxiv}, Mozilla Common Voice 23.0~\cite{Ardila2020CommonVoice}, and Naija Voices~\cite{emezue2025naijavoices}. These datasets, as shown on Table \ref{tab:datasets} span diverse domains, ensuring broad acoustic and linguistic coverage. Dataset selection follows five criteria: (i) inclusion of underrepresented African languages, (ii) availability of validated transcriptions, (iii) compatible licensing, (iv) standardized metadata, and (v) existence of held-out evaluation splits. Rather than maximizing raw data volume, the benchmark emphasizes harmonization of heterogeneous resources into a reproducible and comparable evaluation framework.

\subsection{Pre-processing and Evaluation Splits}
\label{subsec:preprocessing}
All audio is resampled to 16 kHz mono, with transcripts extracted using dataset-specific alignment rules. Text normalization retains punctuation and speaker metadata for analysis, though neither is used in scoring. Before WER and CER calculation, punctuation is removed, text is case-folded, and whitespace is normalized. For datasets lacking predefined splits (e.g., Naija Voices), data is merged across batches and stratified before sampling with $3\%$ of the data selected for evaluation. The total evaluation dataset sums up to $199{,}438$ utterances across $60$ dataset--language splits. Each sample includes a unique ID, audio waveform, transcript, language code, data source, and metadata. Language codes follow the ISO~639-3 standard,\footnote{\url{https://iso639-3.sil.org}} and are provided to models requiring language conditioning using each model's language code inventory. Languages are further grouped into their corresponding language families based on Ethnologue~\cite{ethnologue} to allow metric-level evaluation by language family.

\subsection{Data Reliability by Language}

\begin{figure}[!htb]
\centering
\includegraphics[width=\linewidth]{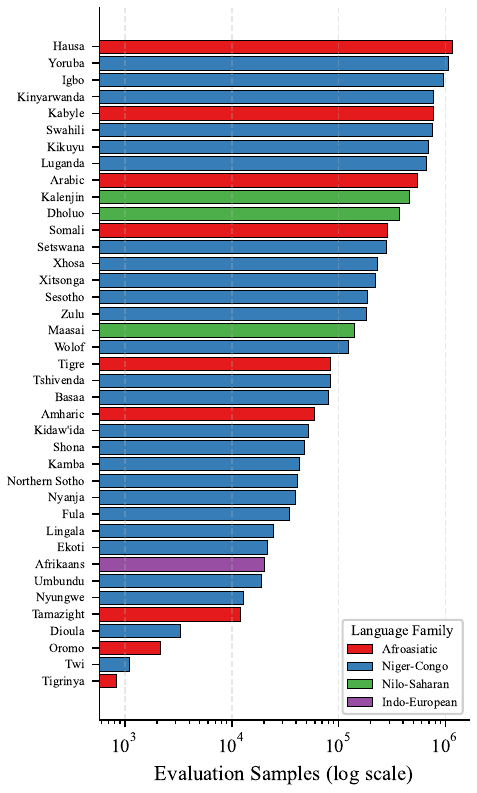}
\caption{Evaluation sample size per language.}
\label{fig:reliability}
\end{figure}

To support robust interpretation of the benchmark results, all evaluations are accompanied by sample count by language as shown in Figure~\ref{fig:reliability}. This makes languages with limited evaluation data readily identifiable and supports uncertainty-aware interpretation of aggregate performance metrics while providing a clearer picture of where additional data availability is most needed.

\subsection{Benchmark Models}

We evaluate 52 publicly available state-of-the-art ASR models spanning 15 model families and three architectural paradigms as shown in Table \ref{tab:all_models}. All models are sourced from open repositories, primarily Hugging Face, ensuring transparency and reproducibility. This diversity allows the benchmark to analyze how model architecture interacts with language coverage, accuracy, and efficiency in low-resource languages.

Although some models are not trained on African languages, all are evaluated under identical cross-lingual conditions, reflecting realistic zero-shot deployment. We confirm that all conclusions remain stable after excluding English-only models, enabling consistent comparison across model families while highlighting strong candidates for downstream adaptation and finetuning.

\subsection{Architectural Paradigms}

To enable principled and architecture-aware comparison, we group evaluated models into three high-level architectural classes based on how speech is encoded, how text is generated, and whether decoding is autoregressive or non-autoregressive. This framework enables controlled analysis of systematic trade-offs between accuracy, efficiency, and robustness that are often conflated in prior ASR benchmarks.

\paragraph{Encoder-decoder}~\cite{chorowski2015attention} ASR models consist of an acoustic encoder paired with an autoregressive text decoder that generates tokens sequentially. This setup enables strong modeling of long-range linguistic dependencies and robust multilingual transfer, but incurs higher inference latency due to sequential decoding.

\paragraph{Encoder-only ASR} models employ a deep acoustic encoder with non-autoregressive decoding objectives, eliminating the need for a text decoder. They typically use alignment-based frameworks such as Connectionist Temporal Classification (CTC), Recurrent Neural Network Transducer (RNN-T), or Token-and-Duration Transducer (TDT)~\cite{Xu2023TDT}. This paradigm supports parallel token prediction and low-latency, high-throughput inference, making it suitable for real-time and large-scale deployment.

\paragraph{LLM-based ASR} treats speech recognition as one task within a multi-task transformer trained on heterogeneous objectives. Speech is encoded into intermediate representations and transcribed via natural-language generation by a large language model. This enables strong contextual reasoning and instruction-following behavior, but typically at higher computational cost than specialized ASR architectures.

Representative models for each paradigm are summarized in Table~\ref{tab:asr_arch_by_column-fixed}. This taxonomy separates architectural design choices from model training scale, highlighting trade-offs often obscured in model-level comparisons.

\begin{table}[!htb]
\centering
\small
\setlength{\tabcolsep}{4pt}
\renewcommand{\arraystretch}{1.3}
\begin{tabular}{lp{5.2cm}}
\toprule
\textbf{Architecture} & \textbf{Model Families} \\
\midrule
Encoder-Only & Wav2Vec2~\cite{Baevski2020Wav2Vec2}, HuBERT~\cite{Hsu2021HuBERT}, Data2Vec~\cite{Baevski2022Data2Vec}, MMS~\cite{pratap2023mms}, OmniASR-CTC~\cite{OmnilingualASR2025} \\
\midrule
Encoder--Decoder & Whisper~\cite{Radford2023Whisper}, Distil-Whisper~\cite{Gandhi2023DistilWhisper}, Lite-Whisper~\cite{Kamahori2025LiteASR}, Moonshine~\cite{Jeffries2024Moonshine}, Parakeet-TDT~\cite{Xu2023TDT}, SeamlessM4T~\cite{Barrault2023SeamlessM4T} \\
\midrule
LLM-Based & Phi-4-MM~\cite{Abouelenin2025Phi4Mini}, Qwen2-Audio~\cite{Chu2024Qwen2Audio}, OmniASR-LLM~\cite{OmnilingualASR2025}, Canary~\cite{Xu2023TDT}, Granite~\cite{Saon2025GraniteSpeech}, Kyutai~STT~\cite{kyutai2025streaming} \\
\bottomrule
\end{tabular}
\caption{ASR model families grouped by architectural paradigm.}
\label{tab:asr_arch_by_column-fixed}
\end{table}

\subsection{Inference Setup, Metrics, and Reproducibility}
All models are evaluated on a single NVIDIA A100-80GB GPU using official inference settings from their model cards, following the Open ASR Leaderboard evaluation framework~\cite{srivastav2025openasrleaderboardreproducible}. To ensure fair efficiency comparisons, audio is processed one utterance at a time (batch size of one) with default model decoding settings unless otherwise specified. 

We report word error rate (WER), character error rate (CER), and inverse real-time factor (RTFx), where higher RTFx indicates faster-than-real-time inference.

\subsection{Statistical Testing Methodology}

Shapiro-Wilk tests reject normality for all groups ($p < 10^{-22}$) and Levene's test confirms unequal variances ($p < 10^{-55}$), ruling out parametric methods. We therefore use the Kruskal-Wallis $H$-test for all group comparisons, with Dunn's post-hoc pairwise tests and Holm-Bonferroni correction~\cite{Demsar2006,Dror2018}. We report Cliff's delta~\cite{Cliff1993} ($\delta$) as a distribution-free pairwise effect size, and Spearman's $\rho$ with bootstrap confidence intervals for correlations.

\section{Results}\label{sec:results}

We report benchmark results across languages, model families, and architectural paradigms, focusing on understanding systematic trade-offs rather than identifying a single best-performing system.

\subsection{Performance by Language}

\begin{figure}[!htb]
\centering
\includegraphics[width=\linewidth]{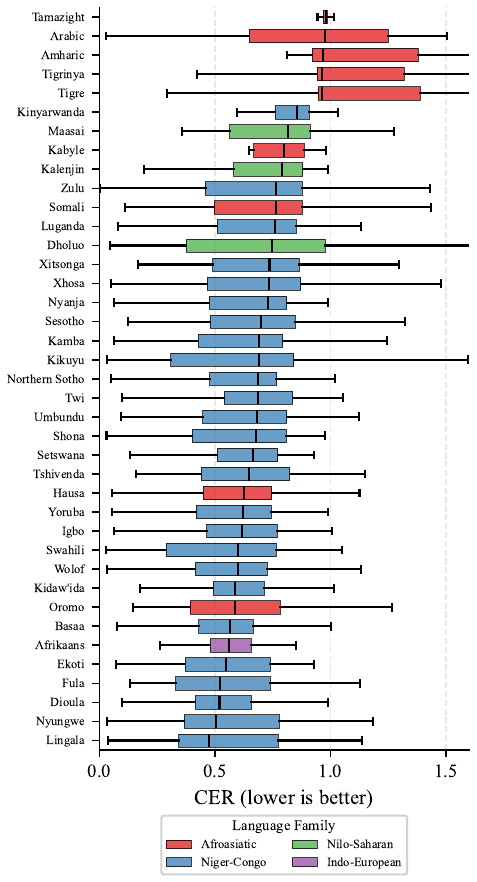}
\caption{CER distribution across the 39 African languages, aggregated over all models.}
\label{fig:language_cer_distribution}
\end{figure}

Figure~\ref{fig:language_cer_distribution} summarizes language-level performance across the 39 evaluated languages. As expected, mid-resource languages (Swahili and Arabic) achieve lower median WER. However, Afroasiatic languages with non-Latin scripts cluster at higher CER, underscoring orthographic challenges for ASR systems trained predominantly on Latin-script languages. Arabic, despite being mid-resource, shows the highest CER ($>$0.90) alongside Amharic and Tamazight exhibiting extreme variance due to script complexity. Lingala, Luganda, and Nyungwe achieve the lowest median CER ($<$0.35).
These results indicate that script and orthographic complexity can outweigh training data availability as primary drivers of ASR performance under cross-lingual conditions.

\subsection{Performance by Model Families}

\begin{figure}[!htb]
\centering
\includegraphics[width=0.9\linewidth]{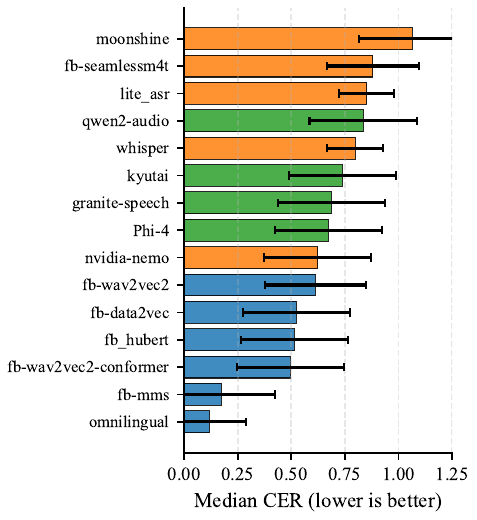}
\caption{Median CER by model family across all evaluated languages.}
\label{fig:model_family_cer}
\end{figure}

Figure~\ref{fig:model_family_cer} presents aggregated CER performance by model family. The omnilingual ASR family achieves the lowest median accuracy error rates across languages given their high language coverage. Within this family, CTC-based variants achieve higher inference efficiency, while LLM-based variants achieve better accuracy. RTFx varies substantially across families based on architecture. The \texttt{nvidia/parakeet-tdt-0.6b-v3} model has the most notable efficiency performance, outperforming all models by more than twice the transcription speed. Table~\ref{tab:top_models} showcases all metrics for the top-performing 15 models.

\begin{table}[!htb]
\centering
\small
\setlength{\tabcolsep}{3pt}
\renewcommand{\arraystretch}{1.1}
\begin{tabular}{p{3.0cm}lrrr}
\toprule
\textbf{Model} & \textbf{Arch.} & \textbf{CER} & \textbf{WER} & \textbf{RTFx} \\
\midrule
omniASR-LLM-7B & LLM & \textbf{0.09} & 0.34 & 7.5 \\
omniASR-LLM-7B\_ZS & LLM & 0.10 & \textbf{0.32} & 5.4 \\
omniASR-LLM-3B & LLM & 0.10 & 0.35 & 8.9 \\
omniASR-LLM-1B & LLM & 0.10 & 0.37 & 9.4 \\
omniASR-CTC-7B & Enc-only & 0.11 & 0.39 & 128.0 \\
omniASR-LLM-300M & LLM & 0.12 & 0.40 & 15.9 \\
omniASR-CTC-3B & Enc-only & 0.12 & 0.43 & 235.4 \\
omniASR-CTC-1B & Enc-only & 0.14 & 0.48 & 383.7 \\
omniASR-CTC-300M & Enc-only & 0.16 & 0.54 & 639.3 \\
mms-1b-all & Enc-only & 0.16 & 0.52 & 303.6 \\
mms-1b-fl102 & Enc-only & 0.20 & 0.52 & 306.9 \\
paza-Phi-4-MM & LLM & 0.46 & 0.96 & 40.4 \\
parakeet-tdt-0.6b-v3 & E-D & 0.48 & 0.97 & \textbf{2181.6} \\
wav2vec2-conformer & Enc-only & 0.49 & 1.11 & 239.8 \\
data2vec-audio-large & Enc-only & 0.50 & 1.04 & 727.3 \\
\bottomrule
\end{tabular}
\caption{Top 15 models ranked by median CER across all languages. Arch.: LLM = LLM-Based, Enc-only = Encoder-only, E-D = Encoder-Decoder.}
\label{tab:top_models}
\end{table}

\subsection{Architectural Trade-offs}

\begin{figure}[!htb]
\centering
\includegraphics[width=0.7\linewidth]{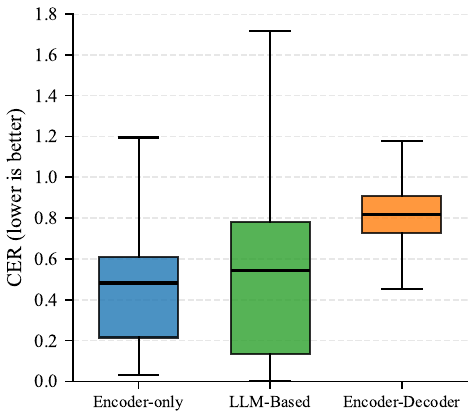}
\caption{CER distribution by architectural paradigm.}
\label{fig:arch_cer}
\end{figure}

\begin{figure}[!htb]
\centering
\includegraphics[width=0.9\linewidth]{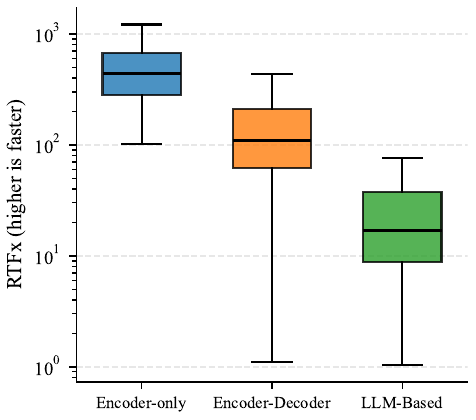}
\caption{RTFx comparison by architecture (log scale).}
\label{fig:architecture_rtfx}
\end{figure}

Figure~\ref{fig:arch_cer} compares CER distributions across architectural paradigms. A Kruskal-Wallis test confirms that CER distributions differ significantly across the three paradigms ($H = 924.8$) with all pairwise comparisons significant after Holm-Bonferroni correction. Encoder-Decoder vs.\ Encoder-only shows the largest pairwise separation (Cliff's $\delta = 0.71$), while Encoder-only vs.\ LLM-Based yields a negligible effect ($\delta = -0.09$), confirming the comparable accuracy between these two paradigms despite fundamentally different decoding strategies. As shown in Figure~\ref{fig:architecture_rtfx} RTFx differences are even more pronounced ($H = 1985.5$), with all pairwise Cliff's $\delta > 0.75$, confirming that efficiency is architecture-driven rather than an artefact of model selection. 

\subsection{Model Performance by Language Family}

\begin{figure}[!htb]
\centering
\includegraphics[width=0.9\linewidth]{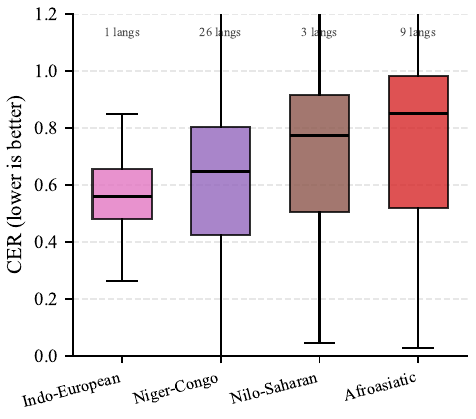}
\caption{CER distribution across language families.}
\label{fig:language_family}
\end{figure}

Table~\ref{tab:language_families} summarizes the distribution of languages across families along with median error rates. Figure~\ref{fig:language_family} shows CER distributions aggregated by language family. Afrikaans, the sole Indo-European language in the benchmark, achieves the lowest error rates, while Afroasiatic and Nilo-Saharan language families exhibit higher median CER and greater variability, likely due to non-Latin scripts and known morphological complexity.

\begin{table}[!htb]
\centering
\small
\setlength{\tabcolsep}{4pt}
\renewcommand{\arraystretch}{1.1}
\begin{tabular}{lrrr}
\toprule
\textbf{Language Family} & \textbf{Langs} & \textbf{CER$_{\text{med}}$} & \textbf{WER$_{\text{med}}$} \\
\midrule
Indo-European & 1 & 0.561 & 0.963 \\
Niger-Congo & 26 & 0.665 & 0.989 \\
Nilo-Saharan & 3 & 0.790 & 0.998 \\
Afroasiatic & 9 & 0.964 & 1.000 \\
\bottomrule
\end{tabular}
\caption{Language families with aggregate accuracy metrics.}
\label{tab:language_families}
\end{table}

\subsection{Language Support}

\begin{figure}[!htb]
\centering
\includegraphics[width=0.9\linewidth]{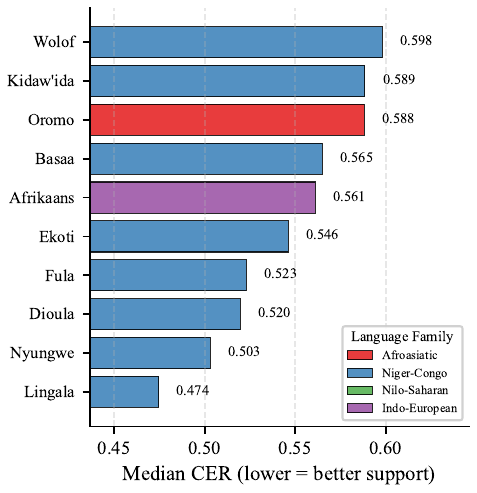}
\caption{Top 10 most supported languages ranked by median CER.}
\label{fig:lang_most}
\end{figure}

\begin{figure}[!htb]
\centering
\includegraphics[width=0.9\linewidth]{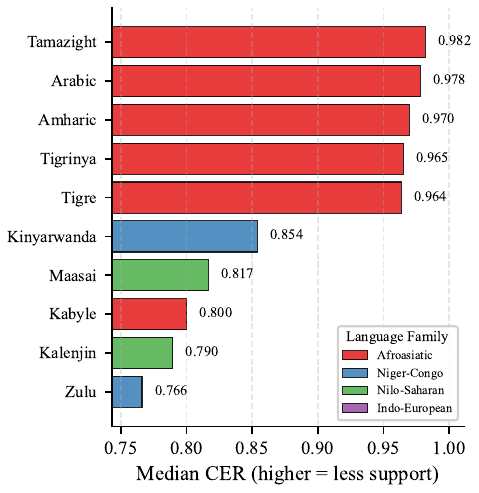}
\caption{Top 10 least supported languages ranked by median CER.}
\label{fig:lang_least}
\end{figure}

We group languages with lower median CER as most supported and those with higher CER as least supported. Figures~\ref{fig:lang_most} and~\ref{fig:lang_least} show the most and least supported languages based on median CER. Niger-Congo languages dominate the top 10 supported languages, with Afrikaans (Indo-European) and Oromo (Afroasiatic) as exceptions. Afroasiatic languages dominate the least supported languages, highlighting the need for script-aware ASR architectures and expanded training data for underrepresented orthographies.

\subsection{Accuracy--Efficiency Trade-offs}

\begin{figure}[!htb]
\centering
\includegraphics[width=\linewidth]{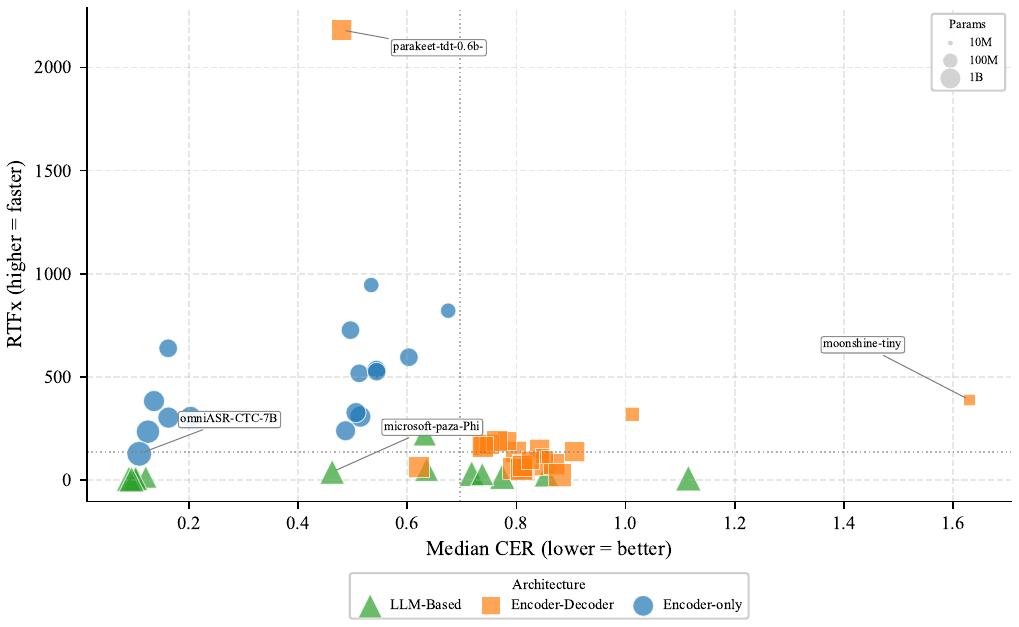}
\caption{Accuracy--efficiency trade-off (CER vs.\ RTFx). Marker color indicates architecture; marker size reflects parameter count.}
\label{fig:accuracy_efficiency}
\end{figure}

Figure~\ref{fig:accuracy_efficiency} illustrates the relationship between transcription accuracy and inference efficiency across all evaluated models. Encoder-only models occupy the high-throughput region, while LLM-based models cluster in the low-error, low-throughput region. Our analysis shows that inference efficiency is primarily architecture‑driven rather than by model size as shown by Figure~\ref{fig:model_size_efficiency}. Language dependence varies by decoding strategy (e.g., CTC models are more speed‑invariant than autoregressive models)

\subsection{Relationship Between WER and CER}

\begin{figure}[!htb]
\centering
\includegraphics[width=\linewidth]{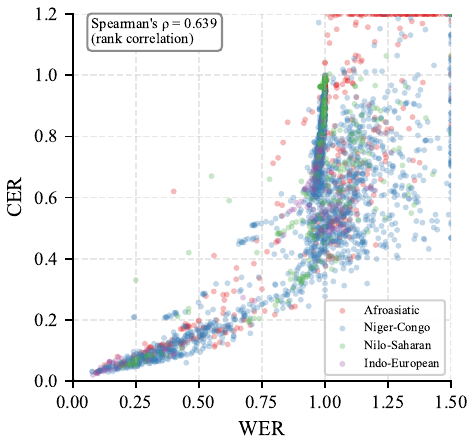}
\caption{WER vs.\ CER correlation across all model--language pairs. Color indicates language family}
\label{fig:wer_cer_correlation}
\end{figure}

Figure~\ref{fig:wer_cer_correlation} shows that WER and CER exhibit a strong positive correlation (Spearman $\rho = 0.64$, 95\% bootstrap CI $[0.61, 0.67]$). Points above the line indicate disproportionately high character-level errors. A Kruskal-Wallis test on CER residuals from a linear fit ($R^2 = 0.67$) confirms significant family-level divergence ($H = 305.3$). Afroasiatic languages show the highest mean positive residual ($+0.14$), confirming that orthographic complexity drives disproportionately high character-level error rates, and that reporting both WER and CER is essential for multilingual ASR evaluation.

\subsection{English-Only vs.\ Multilingual Models}

Of the 52 evaluated models as seen on Table \ref{tab:all_models}, 19 are trained exclusively on English data. To assess whether their inclusion biases conclusions, we repeat all statistical tests after excluding them (removing 1{,}154 of 3{,}168 evaluations, 36.5\%). All conclusions remain significant ($p < 0.0001$) with effect sizes in the same interpretive range. Three effects become \emph{stronger} after removal (architectural CER, WER--CER correlation, model family CER), while two weaken slightly (language family CER, language support gap). Notably, the Encoder-only vs.\ LLM-Based parity dissolves: Cliff's $\delta$ shifts from $-0.09$ (negligible) to $-0.40$ (medium), revealing that apparent parity was partly driven by English-only models achieving moderate CER through English-word overlap rather than genuine multilingual capability.

\section{Discussion}\label{sec:discussion}

\paragraph{Architectural trade-offs.} Statistically significant trade-offs emerge between accuracy and efficiency (Kruskal-Wallis $p < 10^{-200}$, large effect sizes). Encoder-only models offer substantially higher inference efficiency through non-autoregressive decoding while remaining competitive in accuracy where language support is consistent. LLM-based models achieve the lowest median error rates but incur higher computational cost. These differences highlight that architecture, not just parameter size, determines real-world ASR performance.

\paragraph{Accuracy--efficiency trade-off.} Models achieving the lowest error rates generally incur higher computational cost. NVIDIA's parakeet-tdt-0.6b-v3, despite being trained on English-only data, illustrates that architectural optimization can achieve high efficiency without proportional accuracy loss.

\paragraph{Language family differences.} Afroasiatic and Nilo-Saharan languages exhibit higher error rates and greater variability than Niger-Congo languages, consistent with differences in morphology and non-Latin scripts. This suggests that future multilingual ASR systems would benefit from stronger script-aware modelling and tokenization choices, rather than assuming a uniform multilingual decoding setup.

\paragraph{WER vs CER metrics.} Although strongly correlated ($\rho = 0.64$), residual analysis confirms that Afroasiatic languages exhibit systematically higher CER than predicted by their WER ($p < 10^{-65}$). The substantial gap between WER and CER observed for certain languages indicates that character-level evaluation remains more informative for languages with richer morphology or non-Latin script.

\paragraph{English-only model impact.} All benchmark conclusions hold after excluding 19 English-only models. The WER--CER correlation strengthens substantially ($\rho\!: 0.64 \to 0.81$), and the Encoder-only vs.\ LLM-Based parity dissolves ($\delta\!: -0.09 \to -0.40$), suggesting that multilingual coverage, not just architecture, determines cross-lingual accuracy.

\section {Limitations} \label{sec:limitations}
\paragraph{Data leakage.} Public datasets used for evaluation may overlap with pre-training data used by large-scale ASR models introducing potential data leakage. Since model providers do not always disclose full training data composition, benchmark results should be interpreted as relative performance comparisons under shared evaluation conditions rather than absolute generalization estimates.
\paragraph{\textbf{Limited language coverage.}} The benchmark does not capture Africa’s full linguistic diversity. Languages with extremely limited public data remain underrepresented. Future work will prioritize expanding coverage.
\paragraph{Potential language‑model contamination.} LLM‑based ASR models may benefit from textual pretraining overlap which may provide benefit to why LLM-based models display higher accuracy.
\paragraph{Additional analysis.} The benchmark focuses on transcription accuracy and inference efficiency, and does not explicitly evaluate robustness to domain shift, noise, or demographic factors.

\section{Conclusion and Future Work}\label{sec:conclusion}

This benchmark's findings underscore the significance of a large-scale low and mid-resource language ASR benchmark that integrates architectural analysis, deployment-relevant trade-offs on accuracy and efficiency metrics, and linguistically grounded evaluation. Together, they provide practical foundations for inclusive ASR development for diverse low-resource languages.

\textbf{Future Work}. Future work will broaden language coverage to other low-resource regions to enable cross-regional comparison, incorporate speaker and language identification, evaluate controlled fine-tuning and model adaptations, and assess real-world model robustness to noise, code-mixing/switching and domain shifts through data augmentation. We also plan to release inference outputs with detailed insertion, deletion, and substitution error analyses.

By providing a transparent and extensible benchmark, this work lays the groundwork for more principled, deployment-aware multilingual ASR research and continued progress toward inclusive speech technologies.

\appendix

\newpage

\begin{center}
{\Large\textbf{Appendix}}
\end{center}
\vspace{1em}
\section{Evaluation Datasets}
\label{sec:appendix_datasets}

Table~\ref{tab:datasets} summarizes the six publicly available datasets unified to construct the evaluation benchmark, capturing 61 dataset--language test splits across 39 African languages, totaling 199,438 evaluation samples.

\begin{table}[!htb]
\centering
\small
\setlength{\tabcolsep}{3pt}
\renewcommand{\arraystretch}{1.15}
\begin{tabular}{@{}>{\raggedright\arraybackslash}p{3.2cm}>{\raggedright\arraybackslash}p{2.5cm}rr@{}}
\toprule
\textbf{Dataset} & \textbf{Description} & \textbf{Langs} & \textbf{Samples} \\
\midrule
Mozilla Common Voice 23.0 & Crowdsourced speech & 23 & 78,546 \\
Naija Voices & Conversational speech & 3 & 57,529 \\
African Next Voices Kenya & Conversational speech & 5 & 24,508 \\
African Next Voices South Africa & Conversational speech & 6 & 21,692 \\
Google FLEURS & Read speech & 20 & 12,813 \\
ALFFA & Broadcast/read speech & 3 & 4,350 \\
\midrule
\textbf{Total} & & \textbf{39} & \textbf{199,438} \\
\bottomrule
\end{tabular}
\caption{Summary of the six evaluation datasets spanning 39 African languages.}
\label{tab:datasets}
\end{table}

\section{Model Size vs Inference Efficiency}

\begin{figure}[!htb]
\centering
\includegraphics[width=\linewidth]{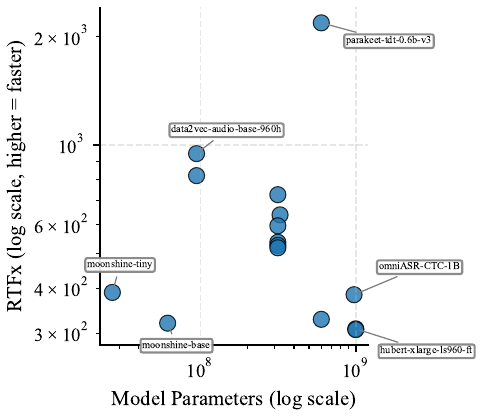}
\caption{Model size vs.\ inference efficiency (RTFx) for the top 15 fastest models.}
\label{fig:model_size_efficiency}
\end{figure}

\section{Languages by Family}
\label{sec:appendix_languages}

Table~\ref{tab:languages_by_family} shows the distribution of languages by family. Niger-Congo is the largest family with 26 languages (67\%), primarily from the Bantu branch. Afroasiatic contributes 9 languages (23\%) spanning Semitic, Cushitic, Chadic, and Berber branches. Nilo-Saharan includes 3 Nilotic languages (8\%), and Indo-European is represented by Afrikaans (3\%).

\begin{table}[!htb]
\centering
\small
\setlength{\tabcolsep}{3pt}
\renewcommand{\arraystretch}{1.05}
\begin{tabular}{@{}l>{\raggedright\arraybackslash}p{2.2cm}>{\raggedright\arraybackslash}p{2.8cm}@{}}
\toprule
\textbf{Family} & \textbf{Language} & \textbf{Sub-family} \\
\midrule
\multicolumn{3}{l}{\textit{Afroasiatic (9 languages)}} \\
& Amharic & Semitic (Ethiopian) \\
& Arabic & Semitic (Central) \\
& Hausa & Chadic (West) \\
& Kabyle & Berber (Northern) \\
& Oromo & Cushitic (Lowland East) \\
& Somali & Cushitic (Lowland East) \\
& Tamazight & Berber (Northern) \\
& Tigre & Semitic (Ethiopian) \\
& Tigrinya & Semitic (Ethiopian) \\
\midrule
\multicolumn{3}{l}{\textit{Indo-European (1 language)}} \\
& Afrikaans & Germanic (West) \\
\midrule
\multicolumn{3}{l}{\textit{Niger-Congo (26 languages)}} \\
& Basaa & Bantu (A.40) \\
& Dioula & Mande (Western) \\
& Ekoti & Bantu (P.30) \\
& Fula & Atlantic (Senegambian) \\
& Igbo & Volta-Niger (Igboid) \\
& Kamba & Bantu (E.70) \\
& Kidaw'ida & Bantu (E.70 Taita) \\
& Kikuyu & Bantu (E.70) \\
& Kinyarwanda & Bantu (J.60) \\
& Lingala & Bantu (C.30) \\
& Luganda & Bantu (J.10) \\
& Northern Sotho & Bantu (S.30) \\
& Nyanja & Bantu (N.30) \\
& Nyungwe & Bantu (P.30) \\
& Sesotho & Bantu (S.30) \\
& Setswana & Bantu (S.30) \\
& Shona & Bantu (S.10) \\
& Swahili & Bantu (G.40) \\
& Tshivenda & Bantu (S.20) \\
& Twi & Volta-Niger (Kwa) \\
& Umbundu & Bantu (R.10) \\
& Wolof & Atlantic (Senegambian) \\
& Xhosa & Bantu (S.40) \\
& Xitsonga & Bantu (S.50) \\
& Yoruba & Volta-Niger (Defoid) \\
& Zulu & Bantu (S.40) \\
\midrule
\multicolumn{3}{l}{\textit{Nilo-Saharan (3 languages)}} \\
& Dholuo & Nilotic (Western) \\
& Kalenjin & Nilotic (Southern) \\
& Maasai & Nilotic (Eastern) \\
\bottomrule
\end{tabular}
\caption{Complete listing of 39 African languages organized by language family.}
\label{tab:languages_by_family}
\end{table}

\section{Full Model Performance Summary}
\label{sec:appendix_models}

Table~\ref{tab:all_models} lists all 52 evaluated models ranked by median CER across languages, with multilingual and English-only models separated. Notably, 33 models are multilingual while 19 are trained exclusively on English data (LibriSpeech 960h checkpoints, Whisper \texttt{.en} variants, Moonshine, Parakeet-TDT, and Kyutai-STT-en). The impact of this composition on benchmark conclusions is analyzed in Section~\ref{sec:results}

\section{Architecture and Model Family Summaries}
\label{sec:appendix_arch}

Table~\ref{tab:architectures} confirms systematic differences across the three major architectural paradigms. Table~\ref{tab:top_rtfx} ranks the top 15 individual models by inference efficiency (RTFx), highlighting the accuracy--efficiency trade-off at the model level

\begin{table}[!htb]
\centering
\small
\setlength{\tabcolsep}{4pt}
\begin{tabular}{lrrrr}
\toprule
\textbf{Architecture} & \textbf{CER$_{\text{med}}$} & \textbf{WER$_{\text{med}}$} & \textbf{RTFx$_{\text{med}}$} & \textbf{N} \\
\midrule
Encoder-only & 0.50 & 1.03 & 450.9 & 16 \\
LLM-Based & 0.63 & 1.04 & 17.4 & 13 \\
Encoder-Decoder & 0.81 & 0.99 & 115.8 & 23 \\
\bottomrule
\end{tabular}
\caption{Architecture-level performance summary (ranked by median CER).}
\label{tab:architectures}
\end{table}

\begin{table}[!htb]
\centering
\small
\setlength{\tabcolsep}{3pt}
\renewcommand{\arraystretch}{1.0}
\begin{tabular}{p{3.45cm}lrrrr}
\toprule
\textbf{Model} & \textbf{Arch.} & \textbf{Size} & \textbf{CER} & \textbf{WER} & \textbf{RTFx} \\
\midrule
parakeet-tdt-0.6b-v3 & E-D & 600M & 0.48 & 0.97 & \textbf{2181.6} \\
data2vec-audio-base-960h & Enc & 94M & 0.53 & 1.09 & 945.7 \\
wav2vec2-base-960h & Enc & 94M & 0.68 & 1.08 & 821.5 \\
data2vec-audio-large-960h & Enc & 315M & 0.50 & 1.04 & 727.3 \\
omniASR-CTC-300M & Enc & 325M & \textbf{0.16} & \textbf{0.54} & 639.3 \\
wav2vec2-large-960h & Enc & 315M & 0.60 & 1.10 & 596.1 \\
wav2vec2-large-960h-lv60 & Enc & 315M & 0.54 & 1.01 & 536.5 \\
wav2vec2-large-robust-ft & Enc & 315M & 0.54 & 1.09 & 526.1 \\
hubert-large-ls960-ft & Enc & 315M & 0.51 & 1.03 & 518.0 \\
moonshine-tiny & E-D & 27M & 1.63 & 2.23 & 389.3 \\
omniASR-CTC-1B & Enc & 975M & 0.14 & 0.48 & 383.7 \\
wav2vec2-conformer-rope & Enc & 600M & 0.51 & 1.08 & 327.9 \\
moonshine-base & E-D & 62M & 1.01 & 1.20 & 319.9 \\
hubert-xlarge-ls960-ft & Enc & 1.0B & 0.51 & 1.03 & 308.5 \\
mms-1b-fl102 & Enc & 1.0B & 0.20 & 0.52 & 306.9 \\
\bottomrule
\end{tabular}
\caption{Top 15 models ranked by inference efficiency (RTFx, higher is faster). Best CER and WER among these models are \textbf{bolded}. Encoder-only models dominate, but omniASR-CTC-300M uniquely combines high efficiency with low error rates.}
\label{tab:top_rtfx}
\end{table}

\begin{table}[!htb]
\centering
\small
\setlength{\tabcolsep}{3pt}
\renewcommand{\arraystretch}{1.0}
\begin{tabular}{p{4.0cm}lrrrr}
\toprule
\textbf{Model} & \textbf{Arch.} & \textbf{Size} & \textbf{CER} & \textbf{WER} & \textbf{RTFx} \\
\midrule
\multicolumn{6}{l}{\textit{Multilingual models (33)}} \\
\midrule
omniASR-LLM-7B & LLM & 7.8B & \textbf{0.09} & 0.34 & 7.5 \\
omniASR-LLM-7B\_ZS & LLM & 7.8B & 0.10 & \textbf{0.32} & 5.4 \\
omniASR-LLM-3B & LLM & 4.4B & 0.10 & 0.35 & 8.9 \\
omniASR-LLM-1B & LLM & 2.3B & 0.10 & 0.37 & 9.4 \\
omniASR-CTC-7B & Enc & 6.5B & 0.11 & 0.39 & 128.0 \\
omniASR-LLM-300M & LLM & 1.6B & 0.12 & 0.40 & 15.9 \\
omniASR-CTC-3B & Enc & 3.1B & 0.12 & 0.43 & 235.4 \\
omniASR-CTC-1B & Enc & 975M & 0.14 & 0.48 & 383.7 \\
omniASR-CTC-300M & Enc & 325M & 0.16 & 0.54 & 639.3 \\
mms-1b-all & Enc & 1.0B & 0.16 & 0.52 & 303.6 \\
mms-1b-fl102 & Enc & 1.0B & 0.20 & 0.52 & 306.9 \\
paza-Phi-4-MM & LLM & 5.6B & 0.46 & 0.96 & 40.4 \\
paza-whisper-large-v3-turbo & E-D & 809M & 0.62 & 0.96 & 62.4 \\
canary-qwen-2.5b & LLM & 2.5B & 0.63 & 1.12 & 222.8 \\
granite-speech-3.3-2b & LLM & 3.0B & 0.64 & 1.09 & 51.9 \\
granite-speech-3.3-8b & LLM & 9.0B & 0.72 & 1.12 & 32.7 \\
whisper-large-v3-turbo & E-D & 809M & 0.74 & 0.97 & 162.1 \\
distil-whisper-large-v3 & E-D & 756M & 0.74 & 0.98 & 168.8 \\
lite-whisper-large-v3-turbo-acc & E-D & 600M & 0.75 & 0.97 & 174.9 \\
canary-1b-v2 & E-D & 1.0B & 0.77 & 1.29 & 189.4 \\
Qwen2-Audio-7B-Instruct & LLM & 8.0B & 0.77 & 1.04 & 17.4 \\
whisper-large & E-D & 1.6B & 0.80 & 0.98 & 56.0 \\
distil-whisper-large-v2 & E-D & 756M & 0.80 & 0.99 & 138.1 \\
whisper-large-v3 & E-D & 1.6B & 0.81 & 0.98 & 53.7 \\
lite-whisper-large-v3-acc & E-D & 1.0B & 0.81 & 0.99 & 71.6 \\
whisper-large-v2 & E-D & 1.6B & 0.81 & 0.98 & 56.3 \\
lite-whisper-large-v3-turbo & E-D & 600M & 0.84 & 0.99 & 149.4 \\
Phi-4-MM & LLM & 5.6B & 0.85 & 1.23 & 27.2 \\
lite-whisper-large-v3 & E-D & 1.0B & 0.85 & 0.99 & 73.0 \\
lite-whisper-large-v3-fast & E-D & 1.0B & 0.87 & 0.99 & 78.0 \\
seamless-m4t-v2-large & E-D & 2.3B & 0.88 & 1.24 & 24.4 \\
lite-whisper-large-v3-turbo-fast & E-D & 600M & 0.91 & 1.00 & 138.4 \\
Qwen2-Audio-7B & LLM & 8.0B & 1.11 & 1.30 & 10.0 \\
\midrule
\multicolumn{6}{l}{\textit{English-only models (19)}} \\
\midrule
parakeet-tdt-0.6b-v3 & E-D & 600M & 0.48 & 0.97 & \textbf{2181.6} \\
wav2vec2-conformer-rel-pos & Enc & 600M & 0.49 & 1.11 & 239.8 \\
data2vec-audio-large-960h & Enc & 315M & 0.50 & 1.04 & 727.3 \\
wav2vec2-conformer-rope-large & Enc & 600M & 0.51 & 1.08 & 327.9 \\
hubert-large-ls960-ft & Enc & 315M & 0.51 & 1.03 & 518.0 \\
hubert-xlarge-ls960-ft & Enc & 1.0B & 0.51 & 1.03 & 308.5 \\
data2vec-audio-base-960h & Enc & 94M & 0.53 & 1.09 & 945.7 \\
wav2vec2-large-960h-lv60 & Enc & 315M & 0.54 & 1.01 & 536.5 \\
wav2vec2-large-robust-ft & Enc & 315M & 0.54 & 1.09 & 526.1 \\
wav2vec2-large-960h & Enc & 315M & 0.60 & 1.10 & 596.1 \\
wav2vec2-base-960h & Enc & 94M & 0.68 & 1.08 & 821.5 \\
kyutai-stt-2.6b-en & LLM & 2.6B & 0.74 & 1.19 & 29.7 \\
distil-whisper-medium.en & E-D & 394M & 0.78 & 0.99 & 188.9 \\
whisper-medium.en & E-D & 769M & 0.81 & 0.99 & 67.8 \\
whisper-small.en & E-D & 244M & 0.82 & 0.99 & 95.7 \\
whisper-base.en & E-D & 73M & 0.85 & 0.99 & 115.8 \\
whisper-tiny.en & E-D & 38M & 0.86 & 1.00 & 111.3 \\
moonshine-base & E-D & 62M & 1.01 & 1.20 & 319.9 \\
moonshine-tiny & E-D & 27M & 1.63 & 2.23 & 389.3 \\
\bottomrule
\end{tabular}
\caption{All 52 evaluated models ranked by median CER within each group. Best CER, WER, and RTFx are \textbf{bolded}. Arch.: LLM = Multitask LLM, Enc = Encoder-only, E-D = Encoder-Decoder.}
\label{tab:all_models}
\end{table}




\bibliographystyle{named}
\bibliography{ijcai26}

\end{document}